\documentclass{article}

\usepackage{microtype}
\usepackage{graphicx}
\usepackage{subfigure}
\usepackage{booktabs} 

\usepackage{hyperref}

\usepackage[accepted]{icml2025}

\usepackage{amsmath}
\usepackage{amssymb}
\usepackage{mathtools}
\usepackage{amsthm}
\usepackage{float}
\usepackage[capitalize,noabbrev]{cleveref}

\theoremstyle{plain}

\theoremstyle{definition}

\theoremstyle{remark}

\usepackage[textsize=tiny]{todonotes}

\icmltitlerunning{GNNs as Predictors of Agentic Workflow Performances}

\begin{document}

\twocolumn[
\icmltitle{GNNs as Predictors of Agentic Workflow Performances}



\icmlsetsymbol{equal}{*}

\begin{icmlauthorlist}
\icmlauthor{Yuanshuo Zhang}{sjtu,bigai,bigai2}
\icmlauthor{Yuchen Hou}{sjtu}
\icmlauthor{Bohan Tang}{ox}
\icmlauthor{Shuo Chen}{bigai,bigai2}
\icmlauthor{Muhan Zhang}{pku}
\icmlauthor{Xiaowen Dong}{ox}
\icmlauthor{Siheng Chen}{sjtu}
\end{icmlauthorlist}

\icmlaffiliation{sjtu}{Shanghai Jiao Tong University}
\icmlaffiliation{bigai}{Beijing Institute for General Artificial Intelligence}
\icmlaffiliation{bigai2}{State Key Laboratory of General Artificial Intelligence, BIGAI}
\icmlaffiliation{ox}{University of Oxford}
\icmlaffiliation{pku}{Peking University}
\icmlcorrespondingauthor{Siheng Chen}{sihengc@sjtu.edu.cn}

\icmlkeywords{Machine Learning, ICML}

\vskip 0.3in
]



\printAffiliationsAndNotice{}  

\begin{abstract}
Agentic workflows invoked by Large Language Models (LLMs) have achieved remarkable success in handling complex tasks. However, optimizing such workflows is costly and inefficient in real-world applications due to extensive invocations of LLMs. To fill this gap, this position paper formulates agentic workflows as computational graphs and advocates Graph Neural Networks (GNNs) as efficient predictors of agentic workflow performances, avoiding repeated LLM invocations for evaluation. To empirically ground this position, we construct \textbf{FLORA-Bench}, a unified platform for benchmarking GNNs for predicting agentic workflow performances. With extensive experiments, we arrive at the following conclusion: GNNs are simple yet effective predictors. This conclusion supports new applications of GNNs and a novel direction towards automating agentic workflow optimization. All codes, models, and data are available at \href{https://github.com/youngsoul0731/Flora-Bench}{this URL}.
\end{abstract}

\setcounter{footnote}{1}
\section{Introduction}
\label{sec:intro}
The rise of Large Language Models (LLMs) and LLM-based agents
have showcased their ability to execute complex tasks across various domains such as code generation~\cite{he2025llm,hu2024self,achiam2023gpt}, problem solving~\cite{wu2024mathchat,xiong2024building}, reasoning~\cite{liu2023dynamic,wu2023autogen,wang2024chain}, and decision making~\cite{qi2024civrealm,huang2024adasociety}.
Building upon these capabilities, agentic workflows, which leverage task decomposition and communications within multi-agent systems, have unlocked the potential to tackle more challenging tasks~\cite{hong2023metagpt,he2025llm,zhang2024aflow,li2023camel}. Notably, OpenAI's five-level framework 
of Artificial General Intelligence (AGI) and Artificial Superhuman Intelligence (ASI) positions such agentic workflows as vital path to transitioning from narrow AI (Level 1-2) to organizational-agentic superintelligence (Level 3-5), where systems autonomously manage large-scale agents. Despite the immense potential of agentic workflows, reliance on manual design limits their ability to achieve the self-evolving nature of AGI and ASI. Therefore, automating agentic workflow optimization is increasingly important as an emerging research topic~\cite{hu2024automated}.

\begin{figure}
    \centering
    \includegraphics[width=0.95\linewidth]{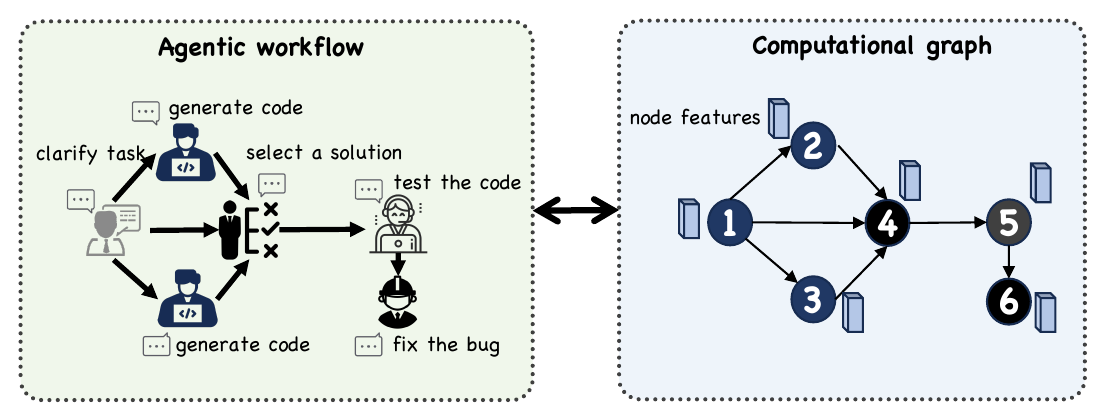}
    \vspace{-1em}
    \caption{An illustration of an agentic workflow and its corresponding computational graph. Nodes are agents handling subtasks and edges are the task dependencies.}
    \label{fig:workflow_demo}
    \vspace{-1em}
\end{figure}

To automate the optimization process, researchers have conceptualized agentic workflows as optimizable graphs, where nodes represent agents handling specific subtasks and edges signify the task dependencies and collaborative interactions as shown in~\cref{fig:workflow_demo}. While existing methods have made strides by optimizing these graphs to improve system performance~\cite{zhugegptswarm, zhang2024g, hu2024self, yuksekgonul2024textgrad, zhang2024aflow,liu2023dynamic}, they share a critical limitation: the optimization process heavily relies on extensive LLM invocations to execute workflows and evaluate their performance. This dependence introduces significant computational, temporal, and financial overhead, making the process costly and inefficient in real-world applications. Therefore, a necessary advancement lies in developing predictors capable of evaluating agentic workflows without costly LLM invocations.

\textbf{In this position paper, we advocate for the adoption of Graph Neural Networks (GNNs) as efficient predictors of agentic workflow performances.} GNNs are powerful and cost-effective tools for capturing graph-level information. Leveraging the message-passing mechanism, GNNs encode the attributes of individual entities (nodes) and the complex relationships (edges) between them, generating representations that encapsulate the overall information of a graph~\cite{errica2019fair,hu2020open,dwivedi2023benchmarking,wu2020comprehensive}. Furthermore, GNNs typically require significantly fewer parameters than LLMs, allowing them to operate efficiently on local hardware. Given these unique properties, GNNs have been utilized to conduct high-throughput screening, which significantly accelerates drug discovery, material design, and biological research by enabling rapid evaluation of molecule properties~\cite{du2024densegnn,jmi.2023.10}. Inspired by these successes in using GNNs for computationally demanding evaluation problems across diverse scientific domains, we propose utilizing GNNs to assess the performance of multi-agent systems without executing them. This approach reduces the considerable computational and financial costs typically incurred during the LLM invocations required by running these systems. By representing agentic workflows as optimizable graphs, as described in the agentic workflow optimization literature, GNNs can predict workflow performances efficiently. These predictions can then be used as rewards to guide the optimization process, enabling the development of more efficient methods for agentic workflow optimization. As such, our position not only addresses the inefficiencies in the existing agentic workflow optimization process but also highlights a novel and impactful application of GNNs in the development of LLM-based multi-agent systems.

However, the feasibility of GNNs as predictors of agentic workflows remains unexplored. To support our position and encourage technological iterations,  we construct the \textbf{FLO}w g\textbf{RA}ph benchmark (\textbf{FLORA-Bench}), a platform for benchmarking GNNs as predictors of agentic workflow performances. Key features of FLORA-Bench include:

$\bullet$ \textbf{Large scale with high quality.} The entire benchmark consists of \textbf{600k} workflow-task pairs and their binary labels denoting success or failure. Moreover, we control the quality of tasks and workflows, as well as inference results by a filtering process. This process ensures the high quality of our data, thereby guaranteeing the reliability and robustness of our experimental conclusions. More details are in~\cref{sec:bench}.

$\bullet$ \textbf{Dual evaluation metrics.} We employ 2 evaluation metrics in our benchmark: i) \textit{accuracy}, which evaluates the correctness of GNN predictions against the ground truth, reflecting the model's ability to assess absolute workflow quality; and ii) \textit{utility}, which assesses the consistency between the workflow rankings predicted by the GNNs and the ground truth rankings, focusing on the model ability in determining the relative quality order of different workflows.

In our experiments, we explore the following research questions to support our position: \textbf{(RQ1)} Can existing GNNs demonstrate promising performance in this prediction task? \textbf{(RQ2)} Are the GNNs' prediction performances robust when LLM driving the agents is different? \textbf{(RQ3)} What about the generalization ability of GNNs across various domains? \textbf{(RQ4)} Whether using GNNs as predictors benefit
the optimization of the agentic workflow?  \textbf{(RQ5)} What are the pros and cons of GNNs over alternative predictors?

By systematically studying these research problems, we make the following key contributions to the community:

$\bullet$ \textbf{Validate the ability of existing GNNs to predict agentic workflow performances.} We show that GNNs are effective and efficient predictors of agentic workflow performances compared to alternative predictors, which serves as an experimental ground for our position and subsequent research.

$\bullet$ \textbf{Find a new application and drive the technological advancements of GNNs.} We explore applying GNNs as predictors of agentic workflow performances, which is an under-explored area. Moreover, while existing GNNs are promising, there is a pressing need to develop GNNs with enhanced generalization capabilities, enabling them to perform effectively across diverse domains. Our FLORA-Bench serves as a fundamental platform for this process. 

$\bullet$ \textbf{Advance an efficient paradigm for agentic workflow optimization.}
We propose a shift in agentic workflow optimization from a trial-and-error approach reliant on repeated LLM invocations to a prediction-driven paradigm powered by rapid evaluations using GNNs. This approach substantially enhances the efficiency of optimization, enabling faster self-evolution of agentic workflows.

\section{Related Work}
\textbf{Agentic Workflows.}
Agentic workflows enable multiple agents to collaborate by sharing information and decomposing complex tasks into manageable subtasks, achieving greater effectiveness compared to single-agent systems \cite{guo2024large}. They have demonstrated notable success in various areas such as code generation~\cite{he2025llm,hu2024self,achiam2023gpt}, math~\cite{zhong2024achieving,wu2024mathchat}, and reasoning~\cite{liu2023dynamic,wu2023autogen,wang2024chain}. Due to the task dependency between nodes, agentic workflows are often modeled as graphs~\cite{zhugegptswarm,zhang2024g,qiao2024benchmarking}  or executable codes~\cite{zhang2024aflow,hu2024automated}.

\textbf{Automated agentic workflow optimization.} Recent works have strived to optimize agentic workflows, which can be categorized into: probability-based and LLM-guided methods. GPTSwarm and G-Designer~\cite{zhugegptswarm,zhang2024g} apply variants of REINFORCE algorithm~\cite{williams1992simple} to optimize the edges and node prompts across a continuous spectrum of probabilistic distributions over a set of feasible Directed Acyclic Graphs (DAGs). 
In contrast, LLM-guided methods refine workflows by directly querying LLMs with workflow structures and the corresponding workflow performances. For instance, ADAS~\cite{hu2024automated} employs LLMs to modify workflows represented in a code-based structure while maintaining historical records as a list; EvoMAC~\cite{hu2024self} uses textual backpropagation informed by environmental feedback to update workflows; and AFLOW~\cite{zhang2024aflow} leverages a Monte Carlo Tree Search (MCTS) variant to enable LLMs to iteratively execute workflows in search of optimal solutions.  Despite their promising results, both types of approaches rely heavily on repeated LLM invocations for workflow execution and performance evaluation, leading to substantial computational, temporal, and financial overhead, which limits their efficiency and practicality in real-world scenarios.

\textbf{Graph Neural Networks.} 
Graph Neural Networks (GNNs) have emerged as a cornerstone in graph representation learning with various architectures~\cite{kipf2016semi,velivckovic2017graph,hamilton2017inductive,chen2024erase}, offering a powerful framework for capturing both local node attributes and global structural patterns through the message-passing mechanism. Foundational studies \cite{errica2019fair, wu2020comprehensive} established that GNNs inherently encode relational dependencies between entities (nodes) and interactions (edges), enabling holistic graph-level representations. Moreover, their parameter efficiency, requiring orders of magnitude fewer parameters than LLMs \cite{hu2020open}, positions GNNs as an efficient solution for resource-constrained settings \cite{dwivedi2023benchmarking}. This efficiency drives their usage in high-throughput applications such as molecular property prediction, where the rapid iteration is critical \cite{du2024densegnn, jmi.2023.10}. Inspired by these successes, we propose for the first time that GNNs are efficient predictors of agentic workflow performances.

\section{Methodology}
In this section, we formalize the key concepts of agentic workflows and our task: agentic workflow performance prediction. Building on these preliminaries, we formally present our proposed method, the workflow graph neural network, which utilizes GNNs as predictors of agentic workflow performances.

\label{sec:preliminaries}
\subsection{Model Agentic Workflows as Graphs}
\textbf{Notations.} In agentic workflows, agents communicate and connect to others through task dependencies. Intuitively, the agentic workflows can be modeled as graphs. Therefore, we formalize the agentic workflow involving $N$ agents as a Directed Acyclic Graph (DAG), $\mathcal{G} = \{\mathcal{V}, \mathcal{E},\mathcal{P}\}$, where $\mathcal{V} = \{v_1, v_2, \dots, v_N\}$ represents the set of agents, with $v_i$ denoting the agent $i$, and $\mathcal{E}$ represents the set of edges defining the connections between agents. Additionally, we define $\mathcal{P} = \{p_1, p_2, \dots, p_N\}$ as the system prompt set, where $p_i$ corresponds to the system prompt of the agent $i$. The agent $i$ receives the task instruction $T$ and outputs from its predecessor agents, which is formulated as:
\begin{equation}
\setlength{\abovedisplayskip}{2pt}
\setlength{\belowdisplayskip}{2pt}
    \mathcal{X}_i = \{T\}\cup\{y_j: v_j\in\mathcal{N}^{\text {(in)}}_i\},
\label{eq:agent_input}
\end{equation}
where $\mathcal{X}_i$ denotes the input of agent $i$, $\mathcal{N}^{\text {(in) }}_i$ denotes the predecessor agents of the agent $i$, and $y_j$ is the output of the agent $j$. For the agent $i$, the output $y_i$ is derived by querying a LLM with the input $x_i$ as follows,
\vspace{-0.2em}
\begin{equation}
\setlength{\abovedisplayskip}{2pt}
\setlength{\belowdisplayskip}{2pt}
    y_i = M(\mathcal{X}_i, p_i),
    \label{eq:agent_output}
\end{equation}
where $M(\cdot)$ denotes the invoked LLM and $p_i$ is the system prompt of agent $i$, describing the subtask assigned to it. 
\textbf{Agentic workflow execution.} Given a task $T$, the agentic workflow $\mathcal{G}$ autonomously executes agents recursively in topological order, as described by~\cref{eq:agent_input,eq:agent_output}. Upon completing the execution of all agents, the response of this agentic workflow is derived as: $r=f_{M}(\mathcal{G},T)$, where $f_M$ is the execution process driven by $M$. Note that due to the sampling in the decoding of $M$, $r$ is random.

\textbf{Agentic workflow evaluation.} Given the agentic workflow's response $r$, we can evaluate its performance. The performance of the agentic workflow $e$ can be obtained as
\begin{equation} 
\setlength{\abovedisplayskip}{2pt}
\setlength{\belowdisplayskip}{2pt}
    e =U(r, T)= U(f_{M}(\mathcal{G},T), T), 
    \label{ground_label}
\end{equation}
where $U(\cdot,\cdot)$ is the criterion function, which is usually implemented through either human evaluation or LLM-as-a-judge. For example, in the current literature of agentic workflow for coding,~\cite{zhugegptswarm,zhang2024g,hu2024automated,hu2024self}, each coding task is associated with human-designed unit tests or the evaluation using GPT-4o. 

According to~\cref{ground_label}, the generation of ground-truth performances involves the complete inference of the agentic workflow and the rigorous evaluation, which is costly with both computational and financial overhead. This poses a huge challenge for the improvement, optimization and iteration of the agentic workflows.



\subsection{Agentic Workflow Performance Prediction Task} 
\label{subsec:task_define}

Here, we present a new direction to evaluate agentic workflows without costly LLM invocations; that is, predicting the performances of agentic workflows through a lightweight model. We detail the model paradigm for this task and the corresponding learning objective.


\begin{figure*}[t!]
    \centering
    \includegraphics[width=0.9\linewidth]{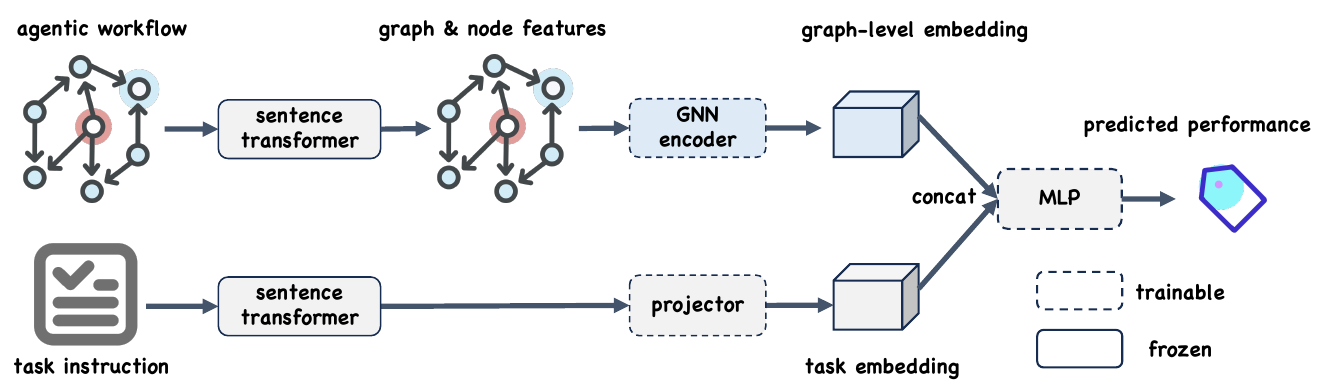}
    \vspace{-1.em}
    \caption{Architecture of workflow graph neural network, a framework for predicting agentic workflow performances with GNNs.}
    \label{fig:flow-gnn}
\vskip -0.1in
\end{figure*}
\textbf{Model paradigm.}  In this task, we require a predictor taking the agentic workflow and the task instruction as inputs to predict the corresponding performance. The predicted performance of the agentic workflow $\mathcal{G}$ on the task instruction $T$ is obtained as
\begin{equation}
\setlength{\abovedisplayskip}{2pt}
\setlength{\belowdisplayskip}{2pt}
    \hat{e} = \mathtt{Preditor}_{\Theta}(\mathcal{G}, T),
\end{equation}
where $\Theta$ denotes the learnable parameters of the predictor. When the predictor is lightweight and precise, it is extremely efficient to use the predictor to guide the optimization of an agentic workflow.



\textbf{Learning objective.} To train this predictor, the learning objective is to ensure the predicted performance $\hat{e}$ closely aligns with the ground-truth performance $e$. Mathematically, the learning objective is then,
\begin{equation}
\setlength{\abovedisplayskip}{2pt}
\setlength{\belowdisplayskip}{2pt}
    \min_{\Theta} \ \mathbb{E}_{(\mathcal{G}, T)} \left[ \mathcal{L}\left( e, \hat{e} \right) \right],
\label{eq:overall_optim}
\end{equation}
where $\mathcal{L}(\cdot,\cdot)$ is a function that quantifies the difference between the ground truth and predicted performance, which could be specifically defined according to the type of prediction such as classification or regression. Importantly, this function measures the performance of $\mathtt{Preditor}_{\Theta}(\cdot,\cdot)$, but not the agentic workflow performance.





\subsection{GNNs as Predictors of Agentic Workflow Performances}
In this section, we propose predicting the performances of the agentic workflows through GNNs; that is, the predicted performance of the agentic workflow $\mathcal{G}$ on the task instruction $T$ is obtained as
$\hat{e}=\mathtt{FLOW\text{-}GNN}_{\Theta}(\mathcal{G},T)$, where $\mathtt{FLOW\text{-}GNN}_{\Theta}(\cdot,\cdot)$ is the workflow graph neural network. The intuition is that the agentic workflow is inherently a computational graph and the graph neural networks can exploit the graph structure effectively and efficiently. The proposed workflow graph neural network involves three stages.


\textbf{Stage 1: Workflow encoding.} This stage utilizes GNNs to encode the given agentic workflow $\mathcal{G}$ into a graph-level embedding for prediction. Given the system prompt $p_i$ of the agent $i$, We first generate the node feature $\boldsymbol{x}_i$:
\begin{equation}
\setlength{\abovedisplayskip}{2pt}
\setlength{\belowdisplayskip}{2pt}
    \boldsymbol{x}_{i} \xleftarrow{} \mathtt{Enc}(p_{i}): v_i \in \mathcal{V},
\end{equation}
where $\mathtt{Enc}(\cdot): p_{i} \mapsto \mathbb{R}^{d_0}$ is a sentence transformer and $d_0$ is the output dimension. After this process, The node features $\boldsymbol{X}=[\boldsymbol{x}_{1},\boldsymbol{x}_{2},\cdots ,\boldsymbol{x}_{N}]^{\top}\in\mathbb{R}^{N\times d_0}$ is derived.  
Given $\boldsymbol{X}$ and $\mathcal{G}$, GNNs generate the graph-level embedding $\boldsymbol{g}$ as,
\begin{equation}
\setlength{\abovedisplayskip}{2pt}
\setlength{\belowdisplayskip}{2pt}
\boldsymbol{g} = \mathtt{GNN}_{\theta}\left(\mathcal{G}, \boldsymbol{X}\right),
    \label{eq:GNN}
\end{equation}
where $\mathtt{GNN}_{\theta}(\cdot,\cdot)$ denotes a GNN parameterized by $\theta$\footnote{Here $\theta$ represents learnable parameters of GNNs, which is different from $\Theta$.}. Specifically, $\mathtt{GNN}_{\theta}(\cdot,\cdot)$ utilizes the graph structure to produce informative node embeddings through iterative processes of message passing and information fusion. The final graph-level representation is then derived by applying a pooling function to these node embeddings \cite{xu2018how}. This detailed process can be formulated as follows,

\vspace{-1em}
\begin{equation}
\setlength{\abovedisplayskip}{2pt}
\setlength{\belowdisplayskip}{2pt}
    \begin{aligned}
                &\boldsymbol{m}_{i}^{(l)} = \mathtt{Aggregate}^{(l)} \left( \left\{ \boldsymbol{h}_{j}^{(l-1)} : v_j \in \mathcal{N}_i^{\text{(in)}} \right\} \right), \\ 
    &\boldsymbol{h}_{i}^{(l)} = \mathtt{Propagate}^{(l)} \left(\boldsymbol{h}_{i}^{(l-1)}, \boldsymbol{m}_{i}^{(l)} \right), \\
    &\boldsymbol{g} = \mathtt{Pool}\left( \{\boldsymbol{h}^{(L)}_{i} \mid v_i \in \mathcal{V} \} \right) , \\
    \label{eq:gnn_h}
\end{aligned}
\end{equation}
\vspace{-2.1em}

where $\boldsymbol{h}_{i}^{(l)}$ is the embeddings of the node $i$ at layer $l$, $\mathcal{N}_i$ is the set of neighbors of node $i$, and $\boldsymbol{m}_{i}^{(l)}$ is the neighborhood embeddings for the node $i$ at layer $l$. The initial node embeddings are set as $\boldsymbol{h}_{i}^{(0)} = \boldsymbol{x}_{i}$. The $\mathtt{Aggregate}^{(l)}(\cdot)$ function performs message passing, aggregating features from the neighbors of node $i$ to generate the neighborhood embeddings $\boldsymbol{m}_{i}^{(l)}$. The $\mathtt{Propagate}^{(l)}(\cdot)$ function updates the embeddings of node $i$ with previous embeddings at layer $l-1$ and the generated neighborhood embeddings at layer $l$. Finally, $\mathtt{Pool}(\cdot)$ represents the pooling operation that takes the node embeddings from the final layer to generate $\boldsymbol{g}$.

 
\textbf{Stage 2: Task encoding.} The task instruction $T$ is encoded into a dense representation $\boldsymbol{t} = \mathtt{Proj}(\mathtt{Enc}(T))$, where $\mathtt{Proj}(\cdot)$ represents a projection module implemented by Multi-Layer Perceptrons (MLPs). This step transforms the raw task instruction into computational embeddings.
 
\textbf{Stage 3: Prediction.} We fuse the graph-level embedding $\boldsymbol{g}$ and the task embedding $\boldsymbol{t}$ through concatenation, yielding the combined embedding $\boldsymbol{o} = \mathtt{Concat}(\boldsymbol{g}, \boldsymbol{t})$, where $\mathtt{Concat}(\cdot,\cdot)$ denotes the concatenation operation. The predicted performance is obtained by, 
\begin{equation}
\setlength{\abovedisplayskip}{2pt}
\setlength{\belowdisplayskip}{2pt}
    \hat{e} = \mathtt{MLP}(\boldsymbol{o}),
\end{equation}
where $\mathtt{MLP}(\cdot)$ denotes a multi-layer perceptron. The implementation of $\mathtt{MLP}(\cdot)$ can be adapted to different types of prediction tasks such as classification and regression, which ensures the generality of the workflow graph neural network.




The architecture of the proposed workflow graph neural networks is summarized in~\cref{fig:flow-gnn}.
\section{FLORA-Bench}
\label{sec:bench}
In this section, we propose the \textbf{FLO}w g\textbf{RA}ph benchmark (\textbf{FLORA-Bench}), an original large-scale benchmark designed to evaluate GNNs as predictors of agentic workflow performances. FLORA-Bench consists of \textbf{600k} workflow-task pairs with annotated performance data across coding, mathematics, and reasoning domains. In the following subsections, we provide a detailed introduction of its construction, key statistics, and evaluation metrics.

\begin{figure}[!]
    \centering
    \includegraphics[width=0.95\linewidth]{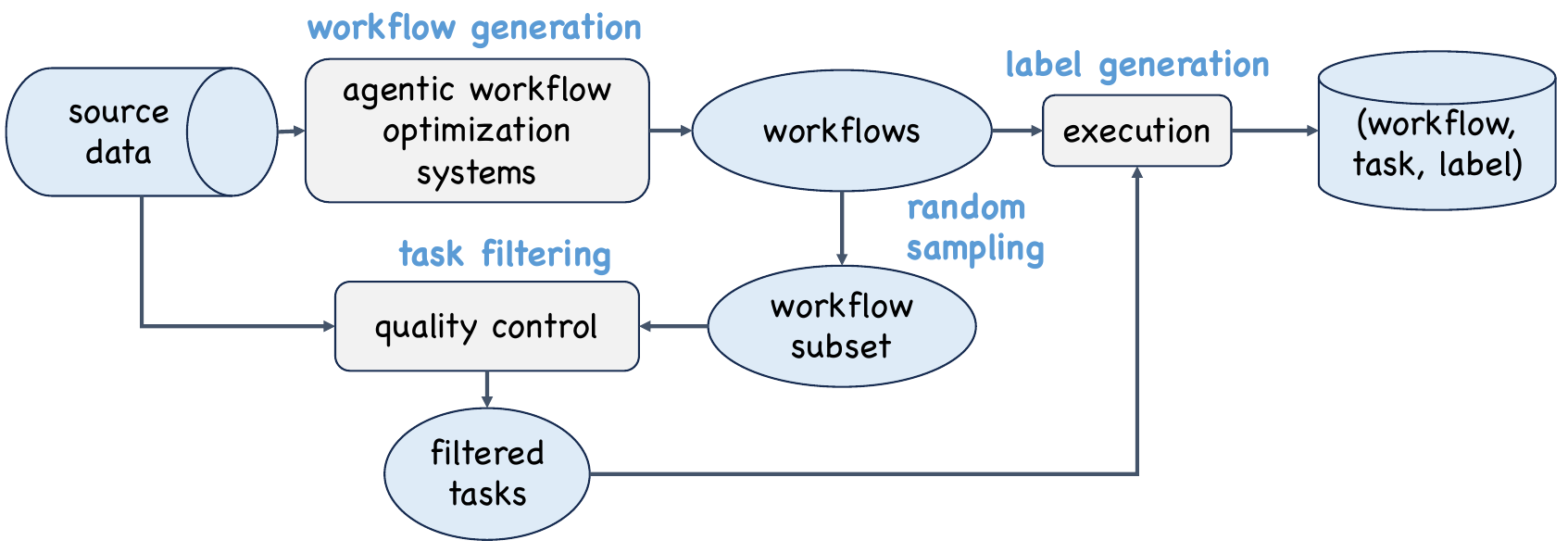}
\vskip -0.1in
    \caption{Pipeline of benchmark construction.}
    \label{fig:pipeline}
\vskip -0.2in
\end{figure}
\subsection{Benchmark Construction}
FLORA-Bench is based on five representative datasets spanning three key topics frequently studied in the agentic workflow literature: coding, mathematics, and reasoning. Specifically, it incorporates the HumanEval~\cite{chen2021evaluating} and MBPP~\cite{austin2021program} for coding, GSM8K~\cite{cobbe2021training} and MATH~\cite{hendrycks2021measuring} for mathematics, and MMLU~\cite{hendrycks2020measuring} for reasoning. We construct FLORA-Bench via a three-stage process: 1) workflow generation, 2) task filtering, and 3) label generation. The construction process is summarized in~\cref{fig:pipeline}.

\begin{figure}[!t]
    \centering
    \includegraphics[width=0.9\columnwidth]{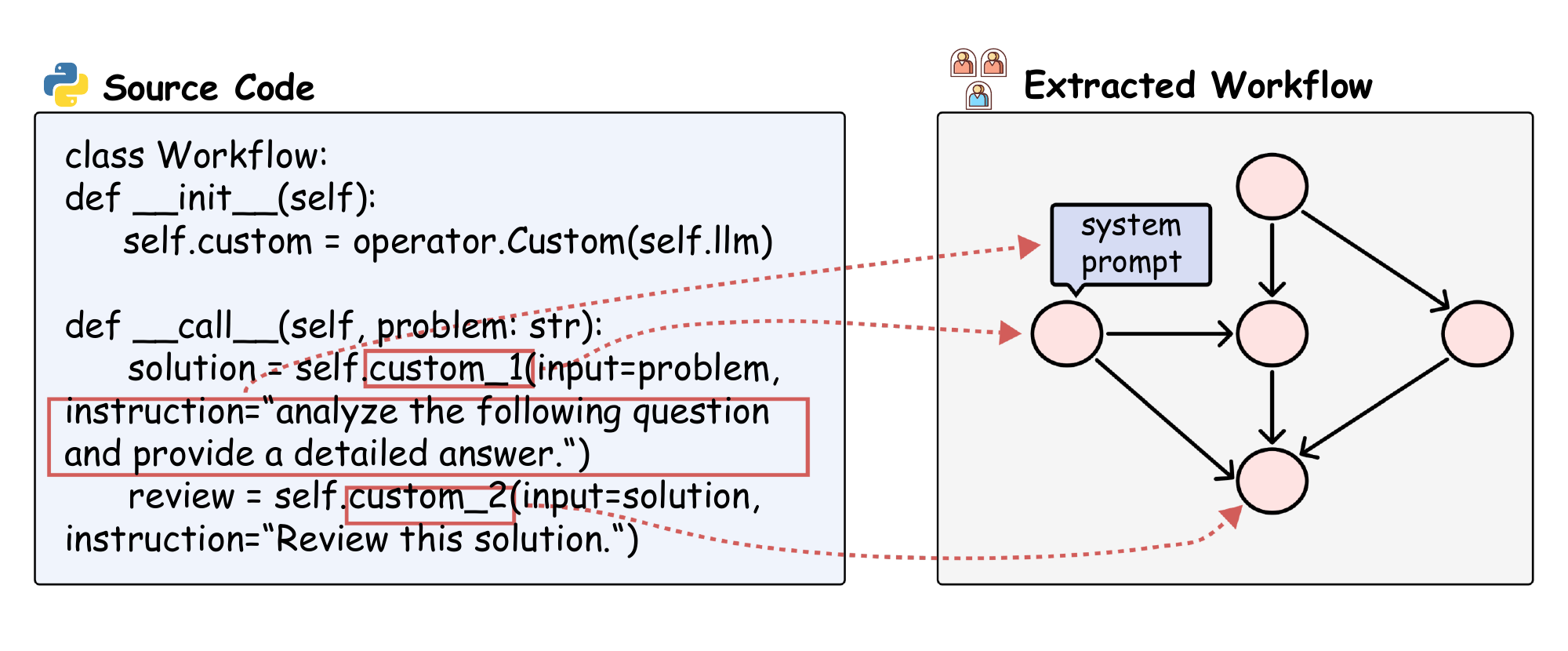}
\vskip -0.2in
    \caption{Workflow Extraction from $AFLOW$ in FLORA-Bench.}
    \label{fig:extarct_workflow_main}
\vskip -0.1in
\end{figure}

\textbf{Workflow generation.} To obtain a diverse and high-quality set of agentic workflows as inputs for the agentic workflow performance prediction task, we derive workflows from the optimization processes of two state-of-the-art agentic workflow optimization systems: $G\text{-}Designer$ (GD) \cite{zhang2024g} and $AFLOW$ (AF) \cite{zhang2024aflow}. We run GD and AF on each task in our selected datasets and extract the agent connections and system prompts from all generated agentic workflows. An extracted agentic workflow is illustrated in~\cref{fig:extarct_workflow_main}, more details of the extraction process are in~\cref{app:workflow_extraction}.


\textbf{Task filtering.} After generating agentic workflows for each task in our selected datasets, we apply a filtering process to ensure the high quality of data in FLORA-Bench. Specifically, we remove tasks that are either too simple or too difficult, as such tasks provide limited value for performance prediction. 
To achieve this, we first infer a domain of workflows on tasks from the selected datasets and compute the success rate for each task. Consequently, we filter out tasks that achieve a success rate higher than $90\%$ or lower than $10\%$. The statistics of the filtered tasks are presented in~\cref{tab:task_filtering}, with further details provided in~\cref{app:task_filtering}.


\begin{table*}[!t]
\centering
\footnotesize
\caption{Overview of 6 domains of our FLORA-Bench. 
}
\label{tab:flora_overview}
\resizebox{1.5\columnwidth}{!}{\begin{tabular}{ccccccc}
\toprule
Domain            & Coding-GD & Coding-AF & Math-GD & Math-AF & Reason-GD & Reason-AF \\ \hline
Num. of workflows & 739       & 38        & 300     & 42      & 189          & 30           \\
Avg. of nodes     & 5.96 & 6.11& 6.06 & 5.49 &5.97&6.58 \\
Num. of tasks     & 233       & 233       & 782     & 782     & 2400         & 2400         \\
Num. of samples   & 30683     & 7362      & 12561   & 4059    & 453600       & 72000        \\ \bottomrule
\end{tabular}}
\vskip -0.1in
\end{table*}

\textbf{Label generation.} After filtering out unsuitable tasks, we get the performance label for each agentic workflow by running inference on the remaining tasks. The output of this inference stage serves as the ground-truth performance, denoted as $e\in\{0,1\}$. To ensure the stability of $e$, we perform inference three times using GPT-4o-mini, setting the temperature to 0 to strictly control randomness in the outputs. The evaluation criterion $U$ is predefined by the source datasets: pass@1 is used for HumanEval and MBPP, while accuracy is used for MATH, GSM8K, and MMLU, as summarized in~\cref{tab:task_filtering}. We denote the final dataset as $\mathcal{D} = \{d_i(\mathcal{G}_i,T_i,e_i)|e_i\in\{0,1\}\}_{i=1}^{|\mathcal{D}|}$. More details about the criterion can be found in~\cref{app:criterion}




\subsection{Benchmark Statistics}
FLORA-Bench consists of six domains: Coding-GD, Coding-AF, Math-GD, Math-AF, Reason-GD, and Reason-AF. Each domain is defined by a task domain (coding, mathematics, or reasoning) and a system domain (GD or AF). For instance, in the Coding-GD domain, the dataset is constructed using agentic workflows generated by the GD system and tasks sourced from the coding domain. An overview of FLORA-Bench is provided in \cref{tab:flora_overview}, illustrating its comprehensive coverage and large scale. To facilitate model training and evaluation, we randomly split each domain into  $\mathcal{D}^{\text{train}}$, $\mathcal{D}^{\text{val}}$, and $\mathcal{D}^{\text{test}}$ according to a $0.8:0.1:0.1$ ratio.



\subsection{Evaluation Metrics}
Our FLORA-Bench employs two evaluation metrics: \textit{accuracy} and \textit{utility}. Accuracy quantifies how well a model predicts agentic workflow performance and is defined as:
\begin{equation}
\setlength{\abovedisplayskip}{2pt}
\setlength{\belowdisplayskip}{2pt}
\label{eq:acc}
    \textit{accuracy} = \frac{1}{\left| \mathcal{D}^{\text{test}}\right|} \sum_{i=1}^{\left| \mathcal{D}^{\text{test}}\right|} \mathbf{1}\left( \hat{e}_i = e_i \right),
\end{equation}
where $\left| \mathcal{D}^{\text{test}}\right|$ is size of test dataset, $\hat{e}_i$ is predicted performance, $e_i$ is ground-truth performance, and $\mathbf{1}(\cdot)$ is the indicator function, which returns 1 if $\hat{e}_{i}=e_{i}$, and 0 otherwise. Utility evaluates the consistency between the workflow rankings predicted by the GNNs and the ground truth rankings, with an emphasis on the model's ability to determine the relative order of different workflows. To obtain utility, we firstly calculate the ground truth success rate  $s_i$ and predicted success rate $\hat{s}_i$ of  $\mathcal{G}_i$ by averaging $e$ and $\hat{e}$ of $\mathcal{G}_i$ on all the tasks in $\mathcal{D}^{\text{test}}$. Then we rank the workflows and extract top-$k$ workflows respectively, according to $s_i$ and $\hat{s}_i$, which produces 2 ordered set of workflows $\mathcal{H} = \{ \mathcal{G}_1, \mathcal{G}_2, \dots, \mathcal{G}_k \}$ and $\hat{\mathcal{H}} = \{ \mathcal{G}_1^{'}, \mathcal{G}_2^{'}, \dots, \mathcal{G}_k^{'}\}$. Finally, we calculate utility by:
\begin{equation}
\setlength{\abovedisplayskip}{2pt}
\setlength{\belowdisplayskip}{2pt}
\label{eq:utility}
    \textit{utility} = \frac{1}{k}\sum_{i=1}^{k}\mathbf{1}(\mathcal{G}_i^{'}\in \mathcal{H}),
\end{equation}
where $\mathbf{1}(\cdot)$ is the indicator function that returns 1 if $\mathcal{G}_i^{'}\in \mathcal{H}$, and 0 otherwise.

\section{Experiments}
We validate our position by answering 5 research questions with extensive experiments on FLORA-Bench: \textbf{(RQ1)} Can existing GNNs demonstrate promising performance in this prediction task? \textbf{(RQ2)} Are the GNNs' prediction performances robust when LLM driving the agents is different? \textbf{(RQ3)} What about the generalization ability of GNNs across various domains? \textbf{(RQ4)} Whether using GNNs as predictors benefit
the optimization of the agentic workflow?  \textbf{(RQ5)} What are the pros and cons of GNNs over alternative predictors such as LLMs?
\subsection{Experimental Setups}
\label{sec:exp_setup}
\textbf{Datasets \& Models.} Our experiments are conducted on our FLORA-Bench across 6 domains mentioned in~\cref{tab:flora_overview}.
 Moreover, we select five GNN-based models:
1) \textbf{GCN}~\cite{kipf2016semi},  
2) \textbf{GAT}~\cite{velivckovic2017graph},  
3) \textbf{GCNII}~\cite{chen2020simple},  
4) \textbf{Graph Transformer}~\cite{shi2020masked} denoted as GT, and  
5) \textbf{One For All}~\cite{liu2023lecheng} denoted as OFA. 
In~\cref{sec:alternative}, we compare these GNN-based models with three alternative predictors:  
1) \textbf{MLP} that processes the raw text-attributed workflow directly using a multilayer perceptron instead of a GNN encoder;  
2) \textbf{FLORA-Bench-Tuned Llama 3B}~\cite{touvron2023llama} denoted as Llama, fine-tuned on the training dataset in 6 domains of FLORA-Bench; and  
3) \textbf{DeepSeek V3}~\cite{liu2024deepseek} denoted as DeepSeek, a mixture-of-experts (MoE) architecture-based LLM, which is prestine.

\textbf{Implementation Details.} For the MLP and GNN baselines, we employ a 2-layer backbone with a hidden dimension of 512. The models are optimized using the Adam optimizer~\cite{kingma2014adam} with a learning rate of $1 \times 10^{-4}$ and a weight decay of $5 \times 10^{-4}$. Training is conducted for 200 epochs on a single NVIDIA RTX-4090 GPU. The best model checkpoint is selected based on the highest \textit{accuracy} on the validation set. For fine-tuning Llama-3.1-8B, we utilize Lora~\cite{DBLP:journals/corr/abs-2106-09685} with the workflow and task text as input, training for 10 epochs. During inference, the temperature is set to 0 to ensure deterministic predictions. For DeepSeek V3, we directly call its API and generate predictions using prompts with temperature also set to 0.

\textbf{Evaluation Metrics.} We use \textit{accuracy} and \textit{utility}, which are defined in~\cref{eq:acc,eq:utility} respectively.

\subsection{RQ1: Can existing GNNs demonstrate promising performances in this prediction task?}
\label{sec:rq1}
The feasibility of GNNs as predictors of agentic workflow performances is the foundation of our position and promotes future research. To validate this, we train and test GNNs on the 6 domains of FLORA-Bench respectively. We evaluate GNNs' performances by $\textit{accuracy}$ and $\textit{utility}$ and report the results in~\cref{tab:main_results}. These results show that all the GNN baselines demonstrate promising prediction \textit{accuracy} and \textit{utility} in most scenarios with an average \textit{accuracy} of \textbf{0.7812} and an average \textit{utility} of \textbf{0.7227} over 6 domains. We also notice that in the Math-GD and Reason-GD domains, all the GNN baselines perform poorly. This may be due to the limited number of unique system prompts in these domains (only 4 and 5, respectively), which likely exacerbates the over-smoothing issue in GNNs. Generally, GNNs have demonstrated significant potential in predicting agentic workflow performances. This supports the feasibility of GNNs as predictors of agentic workflow performances.


\begin{table*}[h]
\centering
\footnotesize
\setlength{\tabcolsep}{1.8pt}
\caption{Main Results of GNN baselines in the test set. GNNs demonstrate promising results in predicting agentic workflow performances.}
\label{tab:main_results}
\begin{tabular}{c|cc|cc|cc|cc|cc|cc}
\toprule
Domain      & \multicolumn{2}{c|}{Coding-GD}    & \multicolumn{2}{c|}{Coding-AF} & \multicolumn{2}{c|}{Math-GD}      & \multicolumn{2}{c|}{Math-AF}      & \multicolumn{2}{c|}{Reason-GD} & \multicolumn{2}{c}{Reason-AF} \\ \hline
Model       & \textit{accuracy}             & \textit{utility}         & \textit{accuracy}             & \textit{utility}      & \textit{accuracy}             & \textit{utility}         & \textit{accuracy}             & \textit{utility}         & \textit{accuracy}             & \textit{utility}      & \textit{accuracy}               & \textit{utility}   \\ \hline
GCN         & 0.8423          & 0.7931          & \textbf{0.8345} & 0.6716       & \underline{0.6412}    & 0.6303          & \textbf{0.8010} & \textbf{0.7491} & 0.7222          & 0.5918       & 0.8674            & 0.8909    \\
GAT         & \underline{0.8514}    & \underline{0.7950}    & 0.8223          & 0.6941       & \textbf{0.6484} & 0.6232          & 0.7985          & \underline{0.7400}    & 0.7216          & 0.5944       & 0.8668            & 0.8966    \\
GCNII       & 0.8381          & 0.7845          &0.8290   & \underline{0.7054} & 0.6356          & \underline{0.6602}    & 0.7936          & 0.7100          & \textbf{0.7229} & 0.5910       & \textbf{0.8708}   & 0.9066    \\
GT          & \textbf{0.8524} & \textbf{0.8020} & 0.8250          & 0.6379       & 0.6325          & 0.6497          & \underline{0.8000}    & 0.7267          & \underline{0.7226}    & \underline{0.6092} & \underline{0.8692}      & 0.8647    \\
OFA & 0.8374          & 0.7593          & \textbf{0.8374} & 0.7593       & 0.6317          & \textbf{0.6665} & 0.7985          & 0.6686          & \textbf{0.7229} & 0.6035       & 0.8142            & 0.9064    \\ \hline
Avg.        & 0.8443          & 0.7868          & 0.8296          & 0.6936       & 0.6379          & 0.6460          & 0.7983          & 0.71889         & 0.7224          & 0.5980       & 0.8577            & 0.8930    \\ \bottomrule
\end{tabular}
\vskip -0.15in
\end{table*}

\begin{table*}[!t]
\centering
\footnotesize
\caption{Results of the cross-system test. GNNs demonstrate a significant decrease in the test domain compared with~\cref{tab:main_results}.}
\label{tab:cross_system}
\resizebox{1.8\columnwidth}{!}{\begin{tabular}{c|cc|cc|cc|cc|cc|cc}
\toprule
       Domain     & \multicolumn{2}{c|}{Coding-GD-AF}                                         & \multicolumn{2}{c|}{Coding-AF-GD}                                         & \multicolumn{2}{c|}{Math-GD-AF}                                           & \multicolumn{2}{c|}{Math-AF-GD}                                           & \multicolumn{2}{c|}{Reason-GD-AF}                                         & \multicolumn{2}{c}{Reason-AF-GD}                                          \\ \hline
         Model   & \textit{accuracy} & \textit{utility} & \textit{accuracy} &\textit{utility} & \textit{accuracy} & \textit{utility} & \textit{accuracy} & \textit{utility} & \textit{accuracy} & \textit{utility} & \textit{accuracy} & \textit{utility} \\ \hline
GCN         & 0.5821          & 0.5733          & 0.5676          & 0.5429          & \textbf{0.6757} & 0.5463          & \underline{0.4964}    & \textbf{0.5192} & 0.5751          & 0.5337          & 0.6137          & \underline{0.5413}    \\
GAT         & 0.5929          & 0.5968          & 0.5725          & 0.5605          & 0.6634          & 0.5270          & 0.4829          & 0.4871          & 0.5607          & 0.5038          & 0.5703          & 0.5312          \\
GCNII       & 0.5875          & \underline{0.6117}    & \textbf{0.6416} & \textbf{0.6267} & \underline{0.6732}    & 0.5296          & 0.4885          & 0.5056          & 0.5593          & 0.5219          & \textbf{0.6555} & 0.5276          \\
GT          & \underline{0.6052}    & \textbf{0.6144} & \underline{0.6083}    & \underline{0.5839}    & 0.5897    & \underline{0.5749}    & 0.4773   & 0.4665  & \underline{0.5650}    & \textbf{0.5486} & 0.5588          & 0.4887          \\
OFA & \textbf{0.6201} & 0.5457          & 0.5897          & 0.5325          & 0.5872          & \textbf{0.6123} & \textbf{0.5060} & \underline{0.5102}    & \textbf{0.5940} & \underline{0.5417}    & \underline{0.6384}    & \textbf{0.5522} \\ \bottomrule
\end{tabular}}
\vskip -0.15in
\end{table*}

\subsection{RQ2: Are the GNNs' prediction performances robust when LLM driving the agents is different?} 
In practical applications,  different LLMs may be used to drive the agentic workflow. We wonder whether  GNNs' prediction performances are robust when LLM driving the agents is different. To validate the robustness of GNNs' performances, which is important for real-world applications, we use 4 LLM: GPT-4o-mini, DeepSeek V3, Qwen 7B, and Mistral 7B, to infer the same agentic workflows and tasks~\footnote{Here we use a subset workflows and tasks in Coding-GD.} collected during benchmark construction to generate 4 datasets. We train and test GNNs on these datasets respectively. We report the  $\textit{accuracy}$ and $\textit{utility}$ in~\cref{tab:llm_consistency}. It can be observed that prediction accuracies of GNNs remain above 0.8 across 4 datasets. Therefore, we believe that GNNs' performances remain robustly strong when LLM driving the agents is different.




\begin{table}[]

\centering
\scriptsize
\caption{ GNNs' prediction performances are robustly strong across 4 datasets generated by different LLMs.}
\label{tab:llm_consistency}
\setlength{\tabcolsep}{1.5pt}
\begin{tabular}{c|cc|cc|cc|cc}
\toprule
                  & \multicolumn{2}{c|}{GPT-4o-mini}  & \multicolumn{2}{c|}{DeepSeek}     & \multicolumn{2}{c|}{Qwen 7B}      & \multicolumn{2}{c}{Mistral 7B}    \\ \cline{2-9} 
                  & \textit{accuracy}             & \textit{utility}         & \textit{accuracy}             & \textit{utility}         & \textit{accuracy}             & \textit{utility}         & \textit{accuracy}             & \textit{utility}         \\ \hline
GCN               &0.8294          & \textbf{0.8040} & \textbf{0.8656} & \textbf{0.7569} &0.8671          & \underline{0.8448}    &0.9258          &0.8848          \\
GAT               & \underline{0.8303}    & \underline{0.8025}    &0.8442          &0.7518          & \textbf{0.8698} &0.8426          & \underline{0.9262}    & \underline{0.8872}    \\
GCNII             &0.8281          &0.7948          &0.8434          & \underline{0.7568}    &0.8517          &0.8271          &0.9094          &0.8589          \\
GT & \textbf{0.8342} &0.7983          & \underline{0.8634}    &0.7306          & \underline{0.8676}    & \textbf{0.8465} & \textbf{0.9280} & \textbf{0.8887} \\
OFA       &0.8124          &0.7192          &0.8473          &0.7323          &0.8451          &0.8042          &0.8913          &0.8506          \\ \bottomrule
\end{tabular}
\vskip -0.15in
\end{table}

\subsection{RQ3: Generalization ability of GNNs to predict agentic workflow performances across domains}

The generalization ability of GNNs to predict agentic workflow performances is crucial in real-world applications. To explore it, we conduct 2 experiments: a cross-system-domain test and a cross-task-domain test. We train GNNs in a domain and test GNNs in another domain. In the cross-system-domain test, the domain for training and the domain for testing differ in the workflow domain but remain the same in the task domain. For example, Coding-GD-AF means we trained GNNs in Coding-GD and tested them in Coding-AF. In the cross-task-domain test, the domain for training and the domain for testing differ in the task domain but remain the same in the workflow domain. For example, Coding-Reason means we trained GNNs in Coding-AF and tested them in the Reason-AF domain. we report $\textit{accuracy}$ and $\textit{utility}$ in~\cref{tab:cross_system,tab:cross_task}.
In these experiments, GNNs demonstrate a significant decrease in the test domain compared with~\cref{tab:main_results} and fail to predict accurately. This indicates that existing GNNs fail to generalize across domains.


\begin{table*}[]
\centering
\footnotesize
\caption{Results of cross-task-domain test. GNNs demonstrate a significant decrease in the test domain compared with~\cref{tab:main_results}.
}
\label{tab:cross_task}
\resizebox{1.9\columnwidth}{!}{\begin{tabular}{c|cc|cc|cc|cc|cc|cc}
\toprule
Domain      & \multicolumn{2}{c|}{Coding-Math}  & \multicolumn{2}{c|}{Coding-Reasoning} & \multicolumn{2}{c|}{Math-Reason} & \multicolumn{2}{c|}{Math-Coding}  & \multicolumn{2}{c|}{Reason-Coding} & \multicolumn{2}{c|}{Reason-Math} \\ \hline
Model       & \textit{accuracy}            & \textit{utility}         & \textit{accuracy}              & \textit{utility}           & \textit{accuracy}             & \textit{utility}          & \textit{accuracy}            & \textit{utility}         & \textit{accuracy}              & \textit{utility}           & \textit{accuracy}             & \textit{utility}          \\ \hline
GCN         & \underline{0.4889}    & 0.5407          & 0.5261            & 0.5329            & \underline{0.4938}     & \underline{0.4669}     & 0.5007          & 0.4875          & 0.3256            & 0.5053            & 0.3342           & 0.5057           \\
GAT         & 0.4595          & 0.4942          & \textbf{0.5371}   & \underline{0.5790}      & 0.4683           & 0.3890           & 0.5102          & 0.4740          & 0.3379            & \textbf{0.5262}   & 0.3342           & 0.5110           \\
GCNII       & \textbf{0.5602} & 0.4449          & \underline{0.5344}      & 0.4593            & \textbf{0.5038}  & \textbf{0.4736}  & 0.3948          & 0.5155          & 0.3813            & 0.5135            & 0.3661           & \textbf{0.5793}  \\
GT          & 0.4767          & 0.5618          & \textbf{0.5371}   & \textbf{0.5795}   & 0.4790           & 0.4363           & \underline{0.5400}    & \underline{0.5620}    & \underline{0.6092}      & \underline{0.5237}      & \textbf{0.4177}  & \underline{0.5291}     \\
OFA & 0.3661          & \textbf{0.6111} & 0.5033            & 0.3982            & \textbf{0.4492}  & 0.4588           & \textbf{0.6540} & \textbf{0.5624} & \textbf{0.6336}   & 0.5060            & \underline{0.3808}     & 0.4527           \\  \bottomrule
\end{tabular}}
\end{table*}

\begin{figure*}
    \centering
    \includegraphics[width=0.95\linewidth]{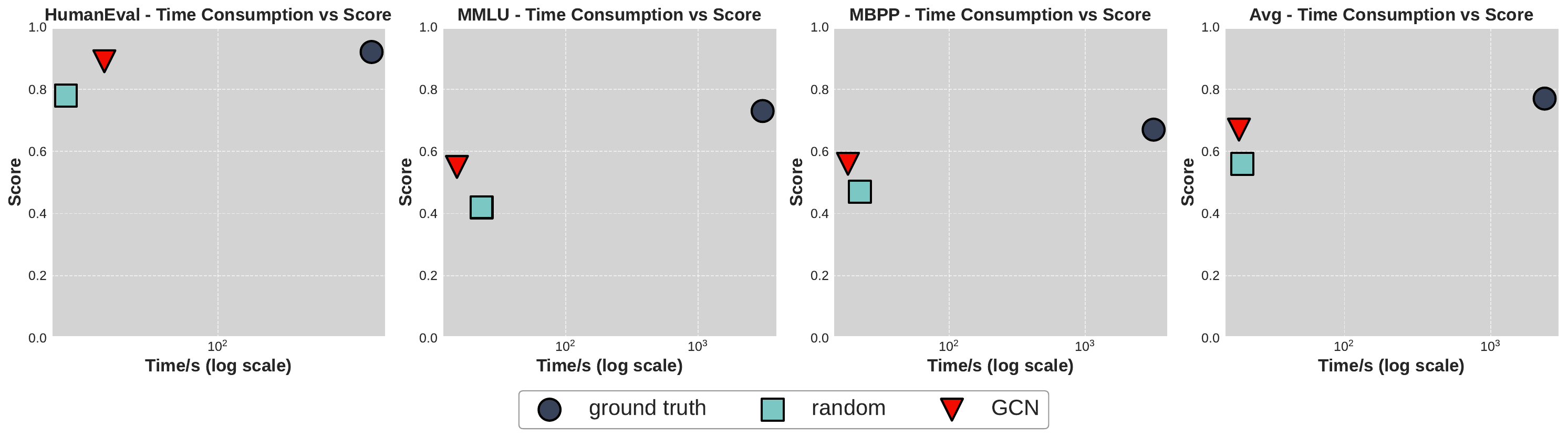}
\vskip -0.2in
    \caption{GCN as the predictor indeed benefits the optimization of the agentic workflow in terms of both effectiveness and efficiency.}
    \label{fig:efficiency_improvement}
\vskip -0.15in
\end{figure*}

\begin{figure}
    \centering
    \includegraphics[width=0.95\linewidth]{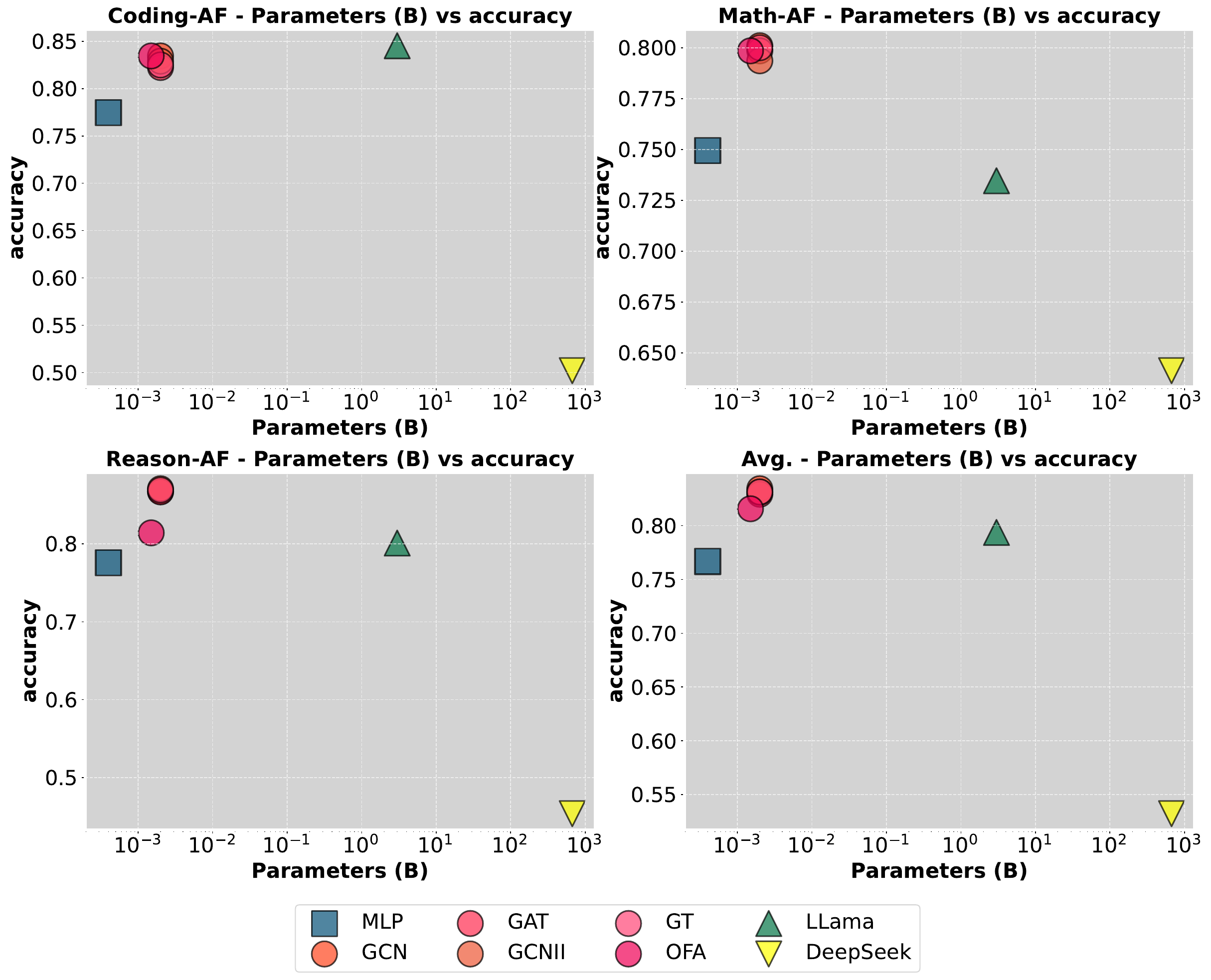}
\vskip -0.15in
    \caption{GNNs are more effective and efficient than others.}
    \label{fig:alternative}
\vskip -0.15in
\end{figure}

\subsection{RQ4: Whether using GNNs as predictors benefits the optimization of the agentic workflow?}
In previous experiments, GNNs have shown feasibility and robustness when predicting agentic workflow performances. However, we may wonder whether they can truly contribute to efficiency improvement and how much performance loss in agentic workflows will be introduced. In this subsection, we examine the improvement of efficiency and loss of agentic workflows when utilizing GNNs as predictors. To make a fair comparison, we use AF as the unified platform to optimize agentic workflows and evaluate the scores of optimized agentic workflows. During the optimization process in a benchmark, a GCN predictor is used to predict agentic workflow performances. The predicted performances are then used as rewards to optimize agentic workflows. When the optimization process is over, the scores of the optimized agentic workflows are calculated by the success rate of all the optimized workflows on test tasks.  We also establish two baselines for comparison: 1) the 'ground truth' baseline, which directly infers the agentic workflow to obtain the ground-truth performances as rewards. and 2) the 'random' baseline, which randomly predicts performances as rewards. We conduct this experiment in 3 benchmarks: HumanEval, MMLU, and MBPP. We report the duration of the optimization process and scores in these 3 benchmarks and their average in~\cref{fig:efficiency_improvement}, where the x-axis is the time consumption(s) and the y-axis is the score of optimized agentic workflow. Detailed results can be found in~\cref{tab:efficiency}. It can be observed that GCN significantly accelerates refinement cycles by 125 times while the performance loss is 0.1 on average. GCN outperforms the 'random' baseline by 0.11 performance gain because GCN correctly predicts performances.  These results support GNNs' potential to advance an efficient paradigm for agentic workflow optimization.

\subsection{RQ5: A comparison with alternative predictors}
\label{sec:alternative}
We compared GNN-based predictors with alternative predictors such as LLMs, as described in~\cref{sec:exp_setup}. We report \textit{accuracy} in~\cref{fig:alternative}. More results are shown in~\cref{tab:alternative}. It indicates that GNNs outperform all alternative predictors in most scenarios. Notably, we observe that DeepSeek-V3, despite its significantly larger parameter volume, fails to achieve accurate predictions compared to GNNs or MLP. This is likely due to the lack of relevant training data for predicting agentic workflow performances and it is not tuned, further highlighting the advantages of using GNNs as predictors. Moreover, the FLORA-Bench-Tuned Llama 3B can outperform GNNs in the Coding-AF domain. However, considering the training cost and inference time, it is not an efficient alternative. Overall, GNNs are more efficient compared to alternatives.

\section{Outlook}

As more efforts are made in utilizing GNNs as predictors of agentic workflow performances, it can bring significant value to the community:

$\bullet$ \textbf{New applications and technological advancements of GNNs.} To make GNNs as predictors of agentic workflow performances applicable to real-world applications, we still need to improve the generalization ability of GNNs to predict agentic workflow performances. This improvement can be approached from two key perspectives. Firstly, there is an urgent need to develop GNN architectures with stronger generalization capability for predicting agentic workflow performances. Secondly, large-scale pretraining of GNNs is essential for real-world applications, where more comprehensive agentic workflows and tasks are required. 

$\bullet$ \textbf{Advancement in an efficient paradigm for agentic workflow optimization.} 
We propose a shift in agentic workflow optimization from a trial-and-error approach reliant on repeated LLM invocations to a prediction-driven paradigm powered by rapid evaluations using GNNs. This approach substantially enhances the efficiency of optimization and can be adapted to different agentic workflow systems, which can bring significant application values.

\section{Conclusions}
In this position paper, we advocate for the adoption of Graph Neural Networks (GNNs) as efficient predictors of agentic workflow performances. To empirically support our position, we make two contributions. Firstly, we formulate agentic workflows, define our task, and propose an architecture FLOW-GNN as the methodological foundation for experiments and future research. Secondly, we construct FLORA-Bench, a large-scale platform for benchmarking GNNs as predictors of agentic workflow performances. Based on this foundation, we conduct extensive experiments and ultimately arrive at the following conclusion: GNNs are simple yet effective predictors. This conclusion supports new applications of GNNs and a novel direction towards automating agentic workflow optimization.


 


\newpage
\nocite{langley00}

\bibliography{example_paper}
\bibliographystyle{icml2025}

\newpage
\appendix
\onecolumn
\section{Details of Benchmark Construction}
\subsection{Workflow Extraction}

In GD system, we directly extract the connections between agents from Directed Acyclic Graphs (DAGs). This approach ensures that the inherent structural relationships within the DAGs are faithfully captured and represented in the workflow.
Moreover, in AF system, we parse the Python source code into an Abstract Syntax Tree (AST). This enables us to capture the information transmission logic of agents through variable propagation. Specifically, each new instance of an agent call is treated as a distinct code unit. When the output variables of one agent instance are used as inputs for the next agent instance, an edge is created between the corresponding nodes in the workflow. Additionally, the specific instructions passed to agents in the source code are integrated with the agent system prompt and embedded as node attributes within the workflow.

\begin{figure}[h]
    \centering                                                      
    \includegraphics[width=0.6\columnwidth]{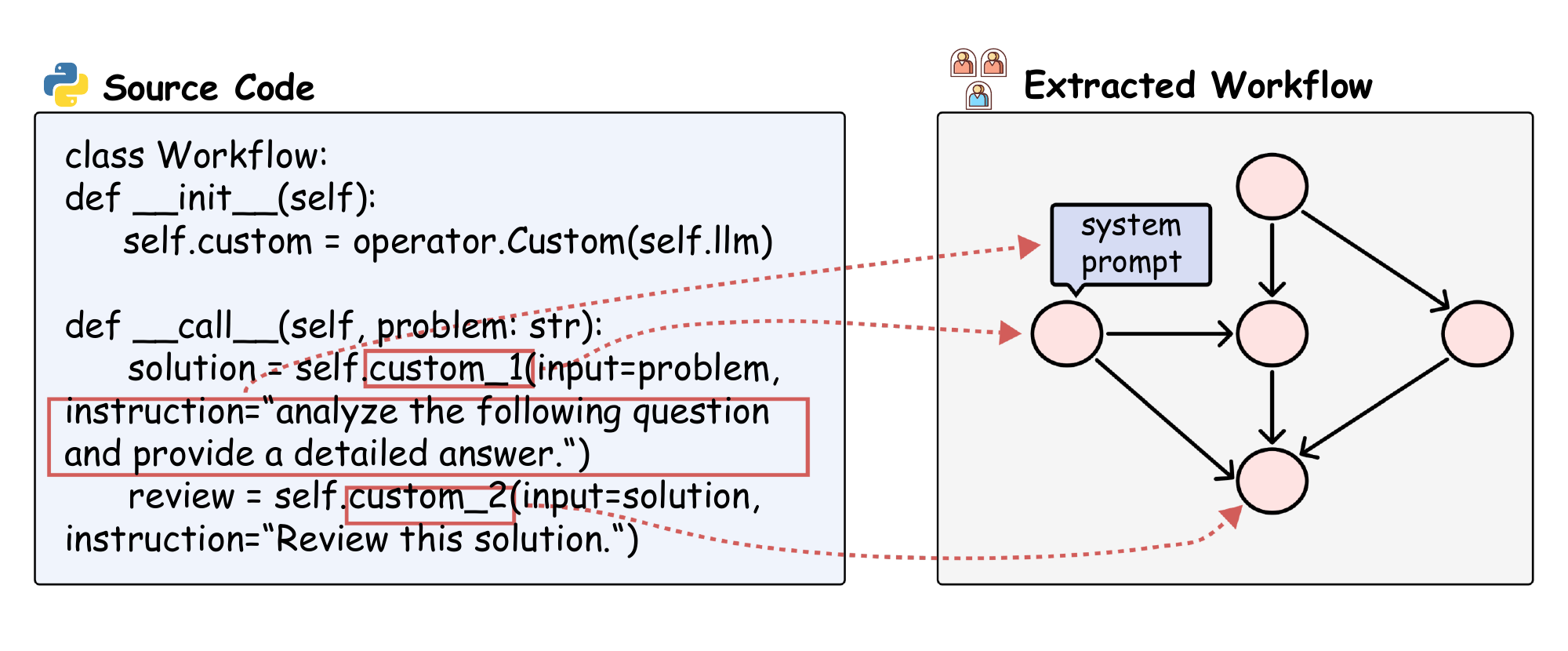}
    \caption{Workflow Extraction from AF in FLORA-Bench.}
    \label{fig:extract_workflow}
\end{figure}

For example, consider the workflow generated from a Python class Workflow in the~ \cref{fig:extract_workflow}. The first node corresponds to the first call to Custom\_1 agent and the second node corresponds to the call to Custom\_2, which reviews the solution generated by Custom\_1. The edges between these nodes represent the flow of the solution from Custom\_1 to Custom\_2. The attributes of the first node include the instruction "analyze the following question and provide a detailed answer".

\label{app:workflow_extraction}
\subsection{Task Filter}
\label{app:task_filtering}
To achieve a more uniform distribution of tasks within our dataset, we implemented a filtering mechanism designed to exclude tasks that are either too complex or too straightforward. Tasks that fall into these categories often provide limited value for training and evaluating predictive models, as they do not offer a balanced knowledge. 

First, We sampled a smaller workflow subset from a specific dataset to capture a wide range of agent profiles. The workflows generated by GD\cite{zhang2024g} system are predefined with specific types and numbers of agents. Therefore, we sampled workflow subsets separately from each unique agent configuration to ensure diversity in agent types and interactions. The workflows generated by AF\cite{zhang2024aflow} system are designed with agent system prompts but feature unpredictable numbers of agents. This is due to the inherent constraints of the optimization framework, specifically the Monte Carlo Tree Search (MCTS) process used by AF, which limits the generation of large-scale workflows. As a result, we included all generated workflows in our sample pool to ensure comprehensive coverage within the scope of workflows that are feasible with the AF. This approach ensures that our dataset encompasses a variety of small-scale workflows, providing a comprehensive view of potential agent interactions.

To achieve balanced performance across tasks, we evaluated all tasks in the current dataset through the workflow subset three times using GPT-4o-mini, with the temperature parameter set to 0. This setup allows for consistent evaluations of task performance.

Then, we calculated the success rate for each task across the workflow subset within the current dataset. Tasks were retained if their success rates fell within a predefined range, as detailed in ~\cref{tab:task_filtering_app}. This range was carefully chosen to avoid including tasks that are either trivially easy (all-success) or intractably difficult (all-fail), thus preserving a challenging yet solvable benchmark. 

For an example, we selected 26 workflows generated by GD for HumanEval\cite{chen2021evaluating} dataset as our subset. These workflows varied in the number of agents, ranging from 4 to 8, and included diverse agent types such as "Programming Expert", "Project Manager", "Test Analyst", "Bug Fixer" and "Algorithm Designer". After evaluating all 161 tasks from the original HumanEval \cite{chen2021evaluating} dataset across this workflow subset, we calculated the success rate for each task. Tasks were retained if their success rates fell within a predefined range of 0.1 to 0.9. This range was chosen to exclude tasks that were either trivially easy (success rate \textgreater 0.9) or intractably difficult (success rate \textless 0.1).  This resulted in a selection of 31 tasks that demonstrated diverse performance across the workflow subset. These tasks were capable of distinguishing performance differences among various workflows, providing a balanced and challenging benchmark for our predictive models.

\begin{table}[t]
\caption{Task filtering  of FLORA-Bench.}
\label{tab:task_filtering_app}
\vskip 0.15in
\begin{center}
\begin{small}
\setlength{\tabcolsep}{2.5pt}
\begin{tabular}{ c c c c c}
\toprule
Data Set & Original Task & Workflow Subset &Success Rate Range & Filtered Task  \\
\midrule
HumanEval  &161 & 26 &[0.1, 0.9] &31  \\
 MBPP   &427& 30& [0.2, 0.8]&202  \\
MATH   &605& 150 &[0.2, 0.8] &178 \\
  GSM8K   &1319& 42&[0.3, 0.9] &404   \\
MMLU &14k+& 20 & [0.25, 0.75]&2400 \\
\bottomrule
\end{tabular}
\end{small}
\end{center}
\vskip -0.1in
\end{table}

\begin{table}[!t]
\caption{Task filtering results of FLORA-Bench.}
\label{tab:task_filtering}
\vskip -0.001in
\begin{center}
\begin{small}
\begin{tabular}{ c c c c}
\toprule
Data Set & Original Task & Filtered Task & Eval Method \\
\midrule
HumanEval  &161&31  &pass@1 \\
 MBPP   &427&202&pass@1  \\
MATH   &605&178& accuracy \\
  GSM8K   &1319&404  & accuracy \\
MMLU &14k+&2400& accuracy \\
\bottomrule
\end{tabular}
\end{small}
\end{center}
\vskip -0.25in
\end{table}

\subsection{Details of Criterion.}
\label{app:criterion}

The methods for generating binary labels vary depending on the nature and requirements of each dataset. Here, we provide a detailed overview of how binary labels are obtained from several key datasets used in our research.
\begin{itemize}
    \item \textbf{HumanEval}\cite{chen2021evaluating} and \textbf{MBPP}\cite{austin2021program}: For these datasets, the primary focus is on evaluating the functionality of the model's output code. Specifically, the code generated by the model is tested to determine whether it can pass predefined unit tests (pass@1). If the code successfully passes the unit tests, it is assigned a binary label of "1" (indicating correctness); otherwise, it receives a binary label of "0" (indicating failure).
    \item \textbf{MATH}\cite{hendrycks2021measuring} and \textbf{GSM8K}\cite{cobbe2021training}: In these datasets, the model's output is evaluated based on the accuracy of the answers. The final answers in the Math dataset are wrapped and delimited with the \texttt{\textbackslash boxed{}} command. For GSM8K, the model's output answer is compared with the ground truth answer within a tolerance level of 1e-6. If the model's answer falls within the acceptable range of the ground truth answer, it is assigned a binary label of "1"; otherwise, it is labeled as "0".
    \item \textbf{MMLU}\cite{hendrycks2020measuring}: The binary label is determined by whether the model's output option matches the correct answer option. Each question in MMLU is a multiple-choice question with four options (A, B, C, and D). The model's performance is evaluated based on its ability to accurately identify the correct answer among the provided choices. If the model selects the correct option, it is labeled as "1"; otherwise, it is labeled as "0".
\end{itemize}
To obtain stable performance, the agentic workflows are inferred 3 times during the label generation in our benchmark construction.
\section{Visualization of Agentic Workflows}
\begin{figure}[htp]
    \centering
    \includegraphics[width=0.75\linewidth]{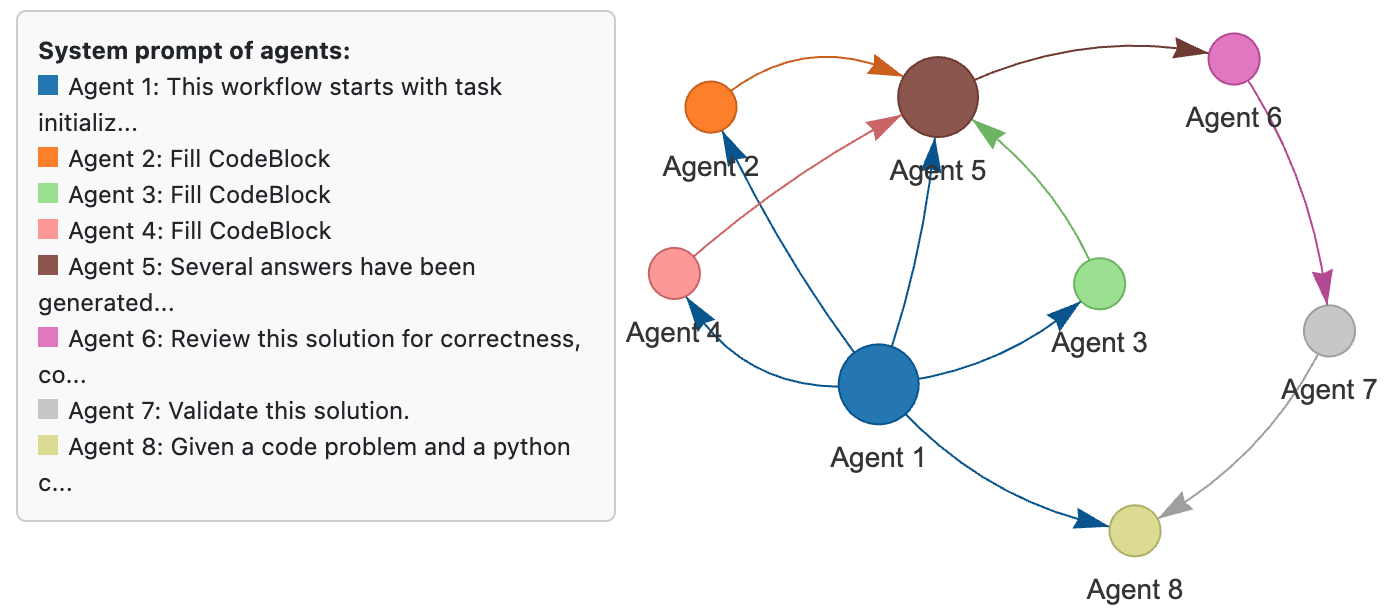}
    \caption{Example of a extracted coding workflow from AF.}
    \label{fig:example_coding}
\end{figure}

In~\cref{fig:example_coding}, the coding workflow starts with task initialization (Agent 1). Then Agent 1 connects to Agents 2, 3, and 4, who each fill different CodeBlocks independently. These components flow into Agent 5, which aggregates the generated answers into a cohesive solution. From there, Agent 6 reviews the solution for correctness before passing it to Agent 7 for final validation. Finally, Agent 8 make final response by receiving the task and the suggestions from Agent 7.

\begin{figure}[htp]
    \centering
    \includegraphics[width=0.75\linewidth]{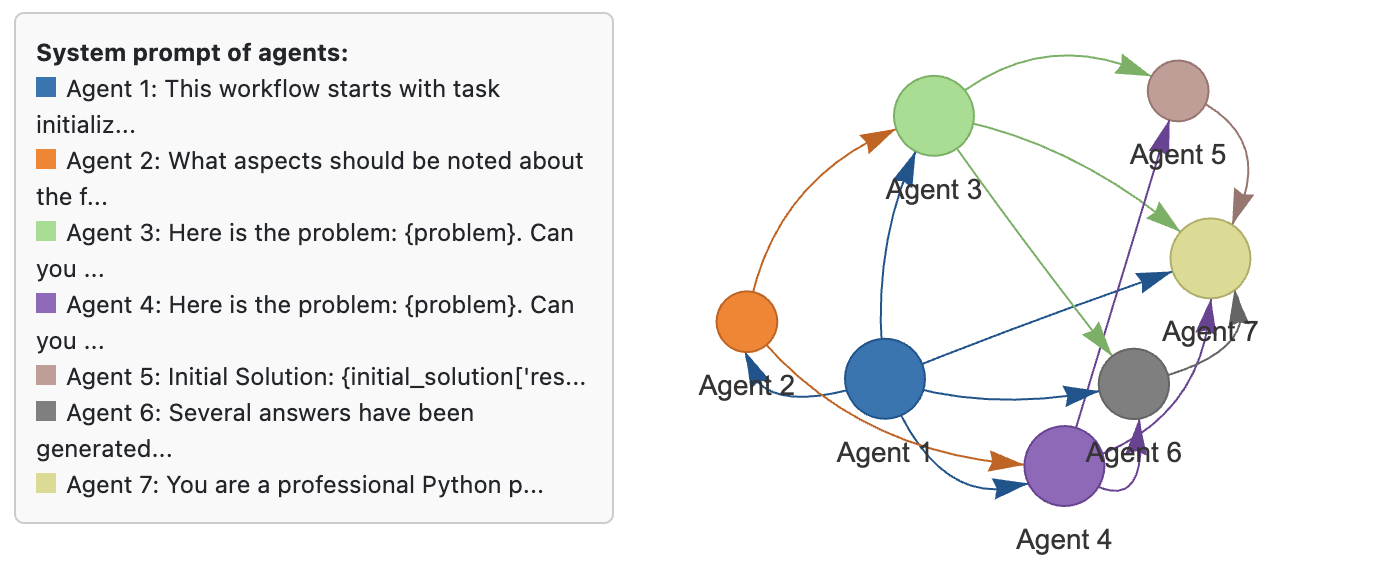}
    \caption{Example of a extracted math workflow from AF.}
    \label{fig:example_math}
\end{figure}

In~\cref{fig:example_math}, the math workflow starts with task initialization (Agent 1). Then Agent 2 make some notes about solving this problem. After this, Agent 3 and 4 receive the notes and attempt to answer this problem. Their answers are passed to Agent 5, which reviews the answers and analyses the difference. Agent 6 receive the problem statement and the answers generated by Agent 3 and 4 to generate additional answer. Finally, the responses of Agent 1, 3, 4, 5, 6 are passed to Agent 7 to make final decision.

\begin{figure}[htp]
    \centering
    \includegraphics[width=0.75\linewidth]{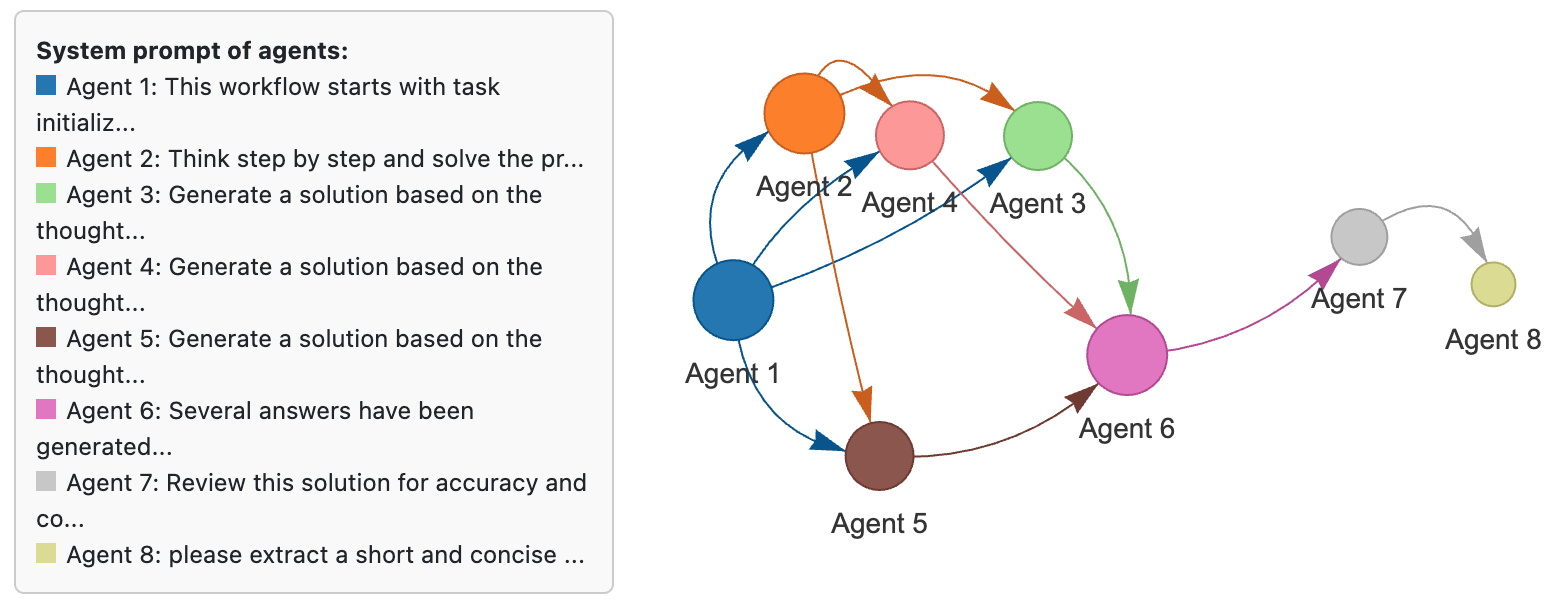}
    \caption{Example of a extracted Reason workflow from AF.}
    \label{fig:example_reason}
\end{figure}

In~\cref{fig:example_reason}, Agent 1 initiates the process and delegates to multiple subsequent agents. Agent 2 serves as the analytical thinker who works step-by-step through the problem and feeds insights to Agents 3, 4, and 5, each of which generates different solution variations based on this thinking. These solutions flow into Agent 6, which aggregates and synthesizes the various answers. The workflow then moves to Agent 7, which reviews the consolidated solution for accuracy and correctness, before concluding with Agent 8, which extracts a short, concise final version. 

\section{Observation of Agentic Workflow Performances}
\begin{figure}[htp]
    \centering
    \includegraphics[width=0.6\linewidth]{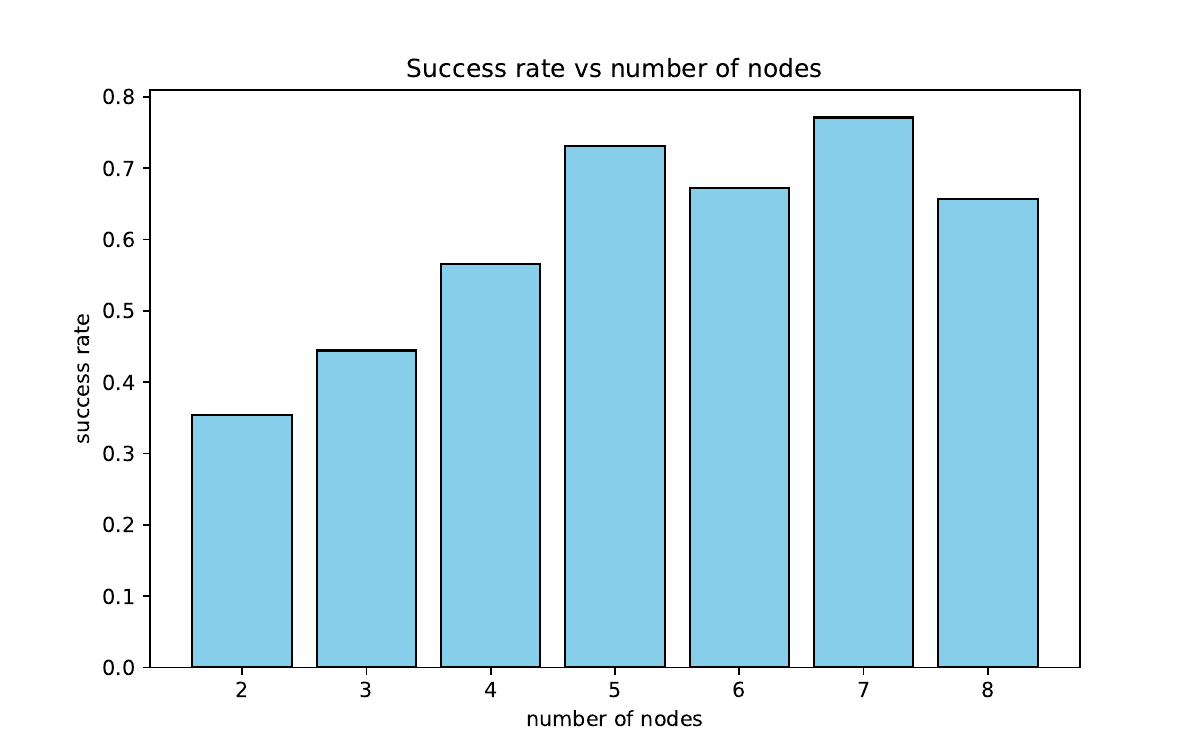}
    \caption{Example of a extracted Reason workflow from AF.}
    \label{fig:num_nodes}
\end{figure}

To analyze the relation between number of nodes and their performances.  We group agentic workflows according to their number of nodes and calculate their average success rate in the collected dataset. The statistics is shown in~\cref{fig:num_nodes}. Results show that when the number of nodes is small, the workflows tend to perform poorly. Within the range from 2 to 5, the success rate increases with the number of nodes. However, as the number of nodes continues to increase, the rate of increase in success rate slows down until it saturates, and the success rate even starts to decrease as the graph complexity increases.

\section{Details of Experiments.}
\subsection{Efficiency Improvement}
We provide more detailed results of~\cref{fig:efficiency_improvement} in~\cref{tab:efficiency}. It can be observed
that GCN significantly accelerates refinement cycles by 125
times while the performance loss is 0.1 in average. And
GCN outperforms the ’random’ baseline by 0.11 performance
gain because GCN correctly predicts performance. These
results support GNNs’ potential to advance an efficient
and effective paradigm for agentic workflow optimization.

\begin{table*}[h]
\centering
\setlength{\tabcolsep}{2.9pt}
\caption{Effieciency improvement (Time/s and Cost/dollar) and performance loss (Score).}
\label{tab:efficiency}
\begin{tabular}{c|ccc|ccc|ccc|ccc}
\toprule
Benchmark     & \multicolumn{3}{c|}{HumanEval} & \multicolumn{3}{c|}{MMLU} & \multicolumn{3}{c|}{MBPP} & \multicolumn{3}{c}{Avg.} \\ \hline
Method        & Score     & Time     & Cost    & Score   & Time   & Cost   & Score   & Time   & Cost   & Score   & Time   & Cost  \\ \hline
ground truth & 0.92      & 728      & 0.13    & 0.73    & 3073   & 6.23   & 0.67    & 3180   & 0.48   & 0.77    & 2327   & 2.28  \\
random        & 0.78      & 14       & 0.01    & 0.42    & 23     & 0.01   & 0.47    & 22     & 0.01   & 0.56    & 20     & 0.01  \\
GCN           & 0.89      & 23       & 0.01    & 0.55    & 15     & 0.01   & 0.56    & 18     & 0.01   & 0.67    & 19     & 0.01  \\ \bottomrule
\end{tabular}
\end{table*}
\subsection{Comparison with Alternative Predictors}
More detailed results of~\cref{fig:alternative} are shown in~\cref{tab:alternative}. It indicates GNNs are simple yet efficient predictors compared to alternatives.

\begin{table}[htp]
\centering
\scriptsize
\setlength{\tabcolsep}{1.0pt}
\caption{Comparison with alternative predictors.}
\label{tab:alternative}
\begin{tabular}{c|cc|cc|cc|cc}
\toprule
          Domain         & \multicolumn{2}{c|}{Coding-AF}       & \multicolumn{2}{c|}{MATH-AF}         & \multicolumn{2}{c|}{Reason-AF}    & \multicolumn{2}{c}{Avg.}          \\ \hline
              Model      & \textit{accuracy}            & \textit{utility}         & \textit{accuracy}           & \textit{utility}         & \textit{accuracy}            & \textit{utility}         & \textit{accuracy}            & \textit{utility}         \\ \hline
MLP                & 0.7748          & 0.6960          & 0.7494          & 0.6888          & 0.7760          & \textbf{0.9299} & 0.7667          & 0.7716          \\ \hline
GCN                & \underline{0.8345}    & 0.6716          & \textbf{0.8010} & \textbf{0.7491} & 0.8674          & 0.8909          & \textbf{0.8343} & 0.7705          \\
GAT                & 0.8223          & 0.6941          & 0.7985          & \underline{0.7400}    & 0.8668          & 0.8966          & 0.8292          & \textbf{0.7769} \\
GCNII              & 0.8290          & \underline{0.7054}    & 0.7936          & 0.7100          & \textbf{0.8708} & 0.9066          & 0.8311          & \underline{0.7740}    \\
GT  & 0.8250          & 0.6379          & \underline{0.8000}    & 0.7267          & \underline{0.8692}    & 0.8647          & \underline{0.8314}    & 0.7431          \\
OFA        & \underline{0.8345}    & \underline{0.7104}    & 0.7985          & 0.6686          & 0.8142          & 0.9064          & 0.8157          & 0.7618          \\ \hline
Llama  & \textbf{0.8453} & \textbf{0.7512} & 0.7346          & 0.6705          & 0.8011          & 0.8360          & 0.7937          & 0.7526          \\
DeepSeek & 0.5020          & 0.4267          & 0.6412          & 0.5097          & 0.4536          & 0.5036          & 0.5323          & 0.4800          \\ \bottomrule
\end{tabular}
\end{table}


\end{document}